# Inertial based Integration with Transformed INS Mechanization in Earth Frame

Lubin Chang, Jingbo Di and Fangjun Qin

*Abstract*—This paper proposes to use a newly-derived transformed inertial navigation system (INS) mechanization to fuse INS with other complementary navigation systems. Through formulating the attitude, velocity and position as one group state of group of double direct spatial isometries $\mathbb{SE}_2(3)$, the transformed INS mechanization has proven to be group affine, which means that the corresponding vector error state model will be trajectory-independent. In order to make use of the transformed INS mechanization in inertial based integration, both the right and left vector error state models are derived. The INS/GPS and INS/Odometer integration are investigated as two representatives of inertial based integration. Some application aspects of the derived error state models in the two applications are presented, which include how to select the error state model, initialization of the $\mathbb{SE}_2(3)$ based error state covariance and feedback correction corresponding to the error state definitions. Extensive Monte Carlo simulations and land vehicle experiments are conducted to evaluate the performance of the derived error state models. It is shown that the most striking superiority of using the derived error state models is their ability to handle the large initial attitude misalignments, which is just the result of log-linearity property of the derived error state models. Therefore, the derived error state models can be used in the so-called attitude alignment for the two applications. Moreover, the derived right error state-space model is also very preferred for long-endurance INS/Odometer integration due to the filtering consistency caused by its less dependence on the global state estimate.

## I. INTRODUCTION

Integrating inertial navigation system (INS) with other complementary navigation systems is a general and popular strategy for accurate and continuous position and navigation [1-6]. The inertial based integration systems can provide attitude, velocity, position information or parts of these information. The extended Kalman filtering (EKF) has proven to be the workhorse for such inertial based integration applications. However, the EKF may suffer from inconsistency problem in some applications, such as Simultaneous Localization And Mapping (SLAM) and visual-inertial navigation systems (VINS) [7-10]. Such inconsistency problem is mainly caused by the false-observability of some state elements. Some researchers have observed that different state error definitions can lead to different performance [11-16]. With such consideration, some researchers make use of alternative state error definitions to circumvent the filtering inconsistency problem [17-22]. Most of these state error definitions fall into the theory of Lie group and Lie algebra. Traditionally, the attitude error is defined on the special orthogonal group $\mathbb{SO}(3)$, while other error state is defined directly in Euclidean space. In [23-25] the group of double direct spatial isometries ($\mathbb{SE}_2(3)$) is introduced to formulate the attitude, velocity and position into a group. With the associated Lie algebra, different velocity and position errors can be derived as opposed to the corresponding errors in Euclidean space. With the vector error corresponding to group error on $\mathbb{SE}_2(3)$, the linearized error state model is trajectory independent and the filtering inconsistency problem can be remedied naturally for SLAM and VINS. With the aforementioned novel error state definition and trajectory-independent state-space model, the integration filter is always termed as invariant EKF (IEKF) [20-23]. Notably, the IEKF has also been successfully applied in for land vehicles and inertial measurement unit (IMU) only integration with vehicle kinematic pseudo-measurement [26, 27].

The trajectory independence of the error state model is rooted in the fact that the group state model satisfies a *group affine* property [23, 28]. Besides the aforementioned ability of remedying the filtering consistency, the *group affine* property also opens door to log-linearity. The log-linearity means that the nonlinear group error can be recovered exactly from the approximated linear vector error. Unfortunately, not every group state model satisfies such *group affine* property. It can be deduced from [20, 21, 26, 27] that only when the earth rotation and Coriolis effect are ignored, the INS mechanization satisfies the *group affine* property. This is because that the earth rotation and Coriolis effect can introduce velocity and position coupling with attitude. However, for high-precision applications, the Earth rotation and Coriolis effect should be taken into account. This is always the case for INS/GPS and INS/Odometer integration [1, 3, 5]. In order to making use of the striking advantage derived by the *group affine* property, an auxiliary velocity is introduced based on the INS mechanization in Earth-Centered Earth-Fixed Frame (ECEF) [24, 25]. The resultant transformed INS mechanization can now satisfy the *group affine* property. However, the transformed INS mechanization is only used for pre-integration on manifolds and the trajectory-independent error state model has not been

The paper was supported in part by National Natural Science Foundation of China (61873275).

The authors are all with the College of Electrical Engineering, Naval University of Engineering. (e-mail: changlubin@163.com, 1452135909@qq.com, haig2005@126.com).

derived. In this paper, we propose to use the transformed INS mechanization for high-precision inertial based integration problems, which necessitates the derivation of the error state model. Through formulating the attitude, transformed velocity and position as an element of $\mathbb{SE}_2(3)$, the error state models corresponding to the transformed INS mechanization with both right and left group error definitions are derived. With INS/GPS and INS/Odometer integration as two representative applications, some application aspects are discussed detailedly and the performance of the derived error state models in the two applications has also been evaluated comprehensively.

The remaining content is organized as follows. Section II presents the general theory corresponding to INS mechanization in ECEF and Lie group. In Section III, the error state models with both right and left error definitions are derived explicitly. Some application aspects have been discussed in order to facilitate the practical applications of the error state models in Section IV. In Section V, the performance of the derived error state models are evaluated using simulation data and the land vehicle field test data for INS/GPS integration. In Section VI, similar evaluations are conducted for INS/Odometer integration. Finally, conclusions are drawn in Section VII.

## II. PROBLEM FORMULATION

Denote the ECEF as $e$ frame, the inertial frame as $i$ frame and the body frame as $b$ frame. INS mechanization in ECEF is given by [29, 30]

$$\dot{\mathbf{C}}_b^e = \mathbf{C}_b^e(\boldsymbol{\omega}_{ib}^b \times) - (\boldsymbol{\omega}_{ie}^e \times)\mathbf{C}_b^e$$
$$\dot{\mathbf{v}}^e = \mathbf{C}_b^e \mathbf{f}^b - 2(\boldsymbol{\omega}_{ie}^e \times)\mathbf{v}^e - (\boldsymbol{\omega}_{ie}^e \times)^2 \mathbf{p}^e + \bar{\mathbf{g}}^e \quad (1)$$
$$\dot{\mathbf{p}}^e = \mathbf{v}^e$$

where $\mathbf{C}_b^e$ denotes the direction cosine matrices (DCM) from body frame to ECEF. $\boldsymbol{\omega}_{ib}^b$ is the body angular rate with respect to inertial frame which can be measured by gyroscopes in body frame. $\boldsymbol{\omega}_{ie}^e$ is the Earth rotation rate expressed in the ECEF. $\mathbf{v}^e = [v_x \; v_y \; v_z]^T$ is the ground velocity expressed in ECEF. $\mathbf{f}^b$ is the specific force measured by accelerometers in the body frame. $\mathbf{p}^e = [x \; y \; z]^T$ is the position vector in ECEF. $\bar{\mathbf{g}}^e$ is the gravitational vector and its relationship with the gravity vector $\mathbf{g}^e$ is given by

$$\mathbf{g}^e = \bar{\mathbf{g}}^e - (\boldsymbol{\omega}_{ie}^e \times)^2 \mathbf{p}^e \quad (2)$$

In [22, 23], an auxiliary velocity vector is introduced as

$$\bar{\mathbf{v}}^e = \mathbf{v}^e + (\boldsymbol{\omega}_{ie}^e \times)\mathbf{p}^e \quad (3)$$

With the introduced auxiliary velocity vector, the INS mechanization is now given by

$$\dot{\mathbf{C}}_b^e = \mathbf{C}_b^e(\boldsymbol{\omega}_{ib}^b \times) - (\boldsymbol{\omega}_{ie}^e \times)\mathbf{C}_b^e$$
$$\dot{\bar{\mathbf{v}}}^e = \mathbf{C}_b^e \mathbf{f}^b - (\boldsymbol{\omega}_{ie}^e \times)\bar{\mathbf{v}}^e + \bar{\mathbf{g}}^e \quad (4)$$
$$\dot{\mathbf{p}}^e = \bar{\mathbf{v}}^e - (\boldsymbol{\omega}_{ie}^e \times)\mathbf{p}^e$$

Formulating the attitude $\mathbf{C}_b^e$, velocity $\bar{\mathbf{v}}^e$ and position $\mathbf{p}^e$ as elements of the group $\mathbb{SE}_2(3)$

$$\chi = \begin{bmatrix} \mathbf{C}_b^e & \bar{\mathbf{v}}^e & \mathbf{p}^e \\ \mathbf{0}_{1\times 3} & 1 & 0 \\ \mathbf{0}_{1\times 3} & 0 & 1 \end{bmatrix} \quad (5)$$

Eq. (4) can now be rewritten in a compact form as
$$\dot{\chi} = f(\chi)$$
$$= \begin{bmatrix} \mathbf{C}_b^e(\boldsymbol{\omega}_{ib}^b \times) - (\boldsymbol{\omega}_{ie}^e \times)\mathbf{C}_b^e & \mathbf{C}_b^e \mathbf{f}^b - (\boldsymbol{\omega}_{ie}^e \times)\bar{\mathbf{v}}^e + \bar{\mathbf{g}}^e & \bar{\mathbf{v}}^e - (\boldsymbol{\omega}_{ie}^e \times)\mathbf{p}^e \\ \mathbf{0}_{1\times 3} & 0 & 0 \\ \mathbf{0}_{1\times 3} & 0 & 0 \end{bmatrix} \quad (6)$$

It can be easily verified that dynamic model (6) satisfies the following relationship

$$f(\chi_1 \chi_2) = f(\chi_1)\chi_2 + \chi_1 f(\chi_2) - \chi_1 f(\mathbf{I}_{5\times 5})\chi_2 \quad (7)$$

where $\chi_1, \chi_2 \in \mathbb{SE}_2(3)$ are the realizations of the group state (5). One dynamic that satisfies such relationship is regarded as *group affine* dynamic [23]. One striking property of the *group affine* dynamic is that its error state model is independent of the global state estimate, which is very profitable when facing observability problem or large initial misalignment in inertial based applications. In next section, we will derive the explicit error state models with both right and left error definitions.

*Remark 1:* Actually, in [31], the same INS mechanization in (4) has been presented. In [31], the INS mechanization is also ingeniously designed to guarantee the resulting group state model is *group affine*. According to [31], the velocity $\bar{\mathbf{v}}^e$ essentially denotes the body velocity with respect to inertial frame, expressed in ECEF. Similarly, $\mathbf{p}^e$ can be viewed as body position with respect to inertial frame, expressed in ECEF. Since the origin of the ECEF is the same with that of the Earth-Centered Inertial frame, $\mathbf{p}^e$ can be viewed as body position with respect to ECEF, expressed in ECEF.

*Remark 2:* The transformed INS mechanization is derived based on the mechanization in ECEF and not the mechanization in local level frame. This is because that there is more coupling between attitude, velocity and position in mechanization in local level frame than that in mechanization in ECEF. Applying similar auxiliary velocity transformation procedure for INS mechanization in local level frame to derive group affine model is an open problem in the future. The newly-derived equivariant filter may be worth considering one research route for such purpose [32, 33].

## III. ERROR STATE MODEL

In this paper, the matrix Lie group is only used to define the nonlinear group error and then to derive vector error according to the relationship between Lie group and Lie algebra. The essential formulas for the corresponding derivations have been presented in Appendix.

### A. Right Error State Model

According to (A14), the right group state error is defined as

$$\boldsymbol{\eta}_r = \chi \tilde{\chi}^{-1} = \begin{bmatrix} \mathbf{C}_b^e \tilde{\mathbf{C}}_b^{eT} & \bar{\mathbf{v}}^e - \mathbf{C}_b^e \tilde{\mathbf{C}}_b^{eT} \tilde{\bar{\mathbf{v}}}^e & \mathbf{p}^e - \mathbf{C}_b^e \tilde{\mathbf{C}}_b^{eT} \tilde{\mathbf{p}}^e \\ \mathbf{0}_{1\times 3} & 1 & 0 \\ \mathbf{0}_{1\times 3} & 0 & 1 \end{bmatrix} \quad (8)$$

where $\tilde{\mathbf{C}}_b^e$, $\tilde{\bar{\mathbf{v}}}^e$ and $\tilde{\mathbf{p}}^e$ are the error-contaminated navigation parameters provided by INS calculation based on INS mechanization (4).

Denote the attitude error corresponding to $\mathbf{C}_b^e \tilde{\mathbf{C}}_b^{eT}$ in Euler angle form as $\boldsymbol{\varphi}_r$. If $\boldsymbol{\varphi}_r$ assumed to be small value, its relationship with $\mathbf{C}_b^e \tilde{\mathbf{C}}_b^{eT}$ is given by

$$\mathbf{C}_b^e \tilde{\mathbf{C}}_b^{eT} \approx \mathbf{I}_{3\times 3} + (\boldsymbol{\varphi}_r \times) \quad (9)$$

With approximations (9) and (A16), the vector-form velocity and position errors corresponding to (8) can be derived as

$$\mathbf{d}\overline{\mathbf{v}}_r = \overline{\mathbf{v}}^e - \mathbf{C}_b^e \tilde{\mathbf{C}}_b^{eT} \tilde{\mathbf{v}}^e \approx (\tilde{\mathbf{v}}^e \times)\boldsymbol{\varphi}_r - \delta\overline{\mathbf{v}}^e \quad (10)$$

$$\mathbf{dp}_r = \mathbf{p}^e - \mathbf{C}_b^e \tilde{\mathbf{C}}_b^{eT} \tilde{\mathbf{p}}^e \approx (\tilde{\mathbf{p}}^e \times)\boldsymbol{\varphi}_r - \delta\mathbf{p}^e \quad (11)$$

where $\delta\overline{\mathbf{v}}^e = \tilde{\mathbf{v}}^e - \overline{\mathbf{v}}^e$ and $\delta\mathbf{p}^e = \tilde{\mathbf{p}}^e - \mathbf{p}^e$.

According to (4), we can easily derive the error model for $\boldsymbol{\varphi}_r$, $\delta\overline{\mathbf{v}}^e$ and $\delta\mathbf{p}^e$ as

$$\dot{\boldsymbol{\varphi}}_r = -(\boldsymbol{\omega}_{ie}^e \times)\boldsymbol{\varphi}_r - \tilde{\mathbf{C}}_b^e \delta\boldsymbol{\omega}_{ib}^b \quad (12a)$$

$$\delta\dot{\overline{\mathbf{v}}}^e = (\tilde{\mathbf{C}}_b^e \tilde{\mathbf{f}}^b \times)\boldsymbol{\varphi}_r - (\boldsymbol{\omega}_{ie}^e \times)\delta\overline{\mathbf{v}}^e + \tilde{\mathbf{C}}_b^e \delta\mathbf{f}^b \quad (12b)$$

$$\delta\dot{\mathbf{p}}^e = \delta\overline{\mathbf{v}}^e - (\boldsymbol{\omega}_{ie}^e \times)\delta\mathbf{p}^e \quad (12c)$$

where $\delta\boldsymbol{\omega}_{ib}^b = \tilde{\boldsymbol{\omega}}_{ib}^b - \boldsymbol{\omega}_{ib}^b$ with $\tilde{\boldsymbol{\omega}}_{ib}^b$ be measured by gyroscopes and $\delta\mathbf{f}^b = \tilde{\mathbf{f}}^b - \mathbf{f}^b$ with $\tilde{\mathbf{f}}^b$ be measured by accelerometers. When deriving (12b), $\overline{\mathbf{g}}^e$ has been assumed to be an error-free constant, although it is also a function of the position. The error state $[\boldsymbol{\varphi}_r^T \ \delta\overline{\mathbf{v}}^{eT} \ \delta\mathbf{p}^{eT}]^T$ can be viewed as definition on $\mathcal{SO}(3) + \mathbb{R}^6$ in contrast with the error state $[\boldsymbol{\varphi}_r^T \ \mathbf{d}\overline{\mathbf{v}}_r^T \ \mathbf{dp}_r^T]^T$ defined on $\mathcal{SE}_2(3)$.

With definitions (10) and (11) and the error state model (12), we now derive the differential equations of $\mathbf{d}\overline{\mathbf{v}}_r$ and $\mathbf{dp}_r$ as follows.

Taking differential operation on both sides of (10) gives

$$\mathbf{d}\dot{\overline{\mathbf{v}}}_r = (\dot{\tilde{\mathbf{v}}}^e \times)\boldsymbol{\varphi}_r + (\tilde{\mathbf{v}}^e \times)\dot{\boldsymbol{\varphi}}_r - \delta\dot{\overline{\mathbf{v}}}^e \quad (13)$$

Substituting (4) and (12) into (13) gives

$$\mathbf{d}\dot{\overline{\mathbf{v}}}_r = (\overline{\mathbf{g}}^e \times)\boldsymbol{\varphi}_r - (\boldsymbol{\omega}_{ie}^e \times)\mathbf{d}\overline{\mathbf{v}}_r - (\tilde{\mathbf{v}}^e \times)\tilde{\mathbf{C}}_b^e \delta\boldsymbol{\omega}_{ib}^b - \tilde{\mathbf{C}}_b^e \delta\mathbf{f}^b \quad (14)$$

Taking differential operation on both sides of (11) gives

$$\mathbf{d}\dot{\mathbf{p}}_r = (\dot{\tilde{\mathbf{p}}}^e \times)\boldsymbol{\varphi}_r + (\tilde{\mathbf{p}}^e \times)\dot{\boldsymbol{\varphi}}_r - \delta\dot{\mathbf{p}}^e \quad (15)$$

Substituting (4) and (12) into (15) gives

$$\begin{aligned}
\mathbf{d}\dot{\mathbf{p}}_r &= \left[(\tilde{\mathbf{v}}^e \times) - (\boldsymbol{\omega}_{ie}^e \times)(\tilde{\mathbf{p}}^e \times)\right]\boldsymbol{\varphi}_r \\
&\quad - (\tilde{\mathbf{p}}^e \times)\tilde{\mathbf{C}}_b^e \delta\boldsymbol{\omega}_{ib}^b - \delta\overline{\mathbf{v}}^e + (\boldsymbol{\omega}_{ie}^e \times)\delta\mathbf{p}^e \\
&= (\tilde{\mathbf{v}}^e \times)\boldsymbol{\varphi}_r - \delta\overline{\mathbf{v}}^e - (\boldsymbol{\omega}_{ie}^e \times)\mathbf{dp}_r - (\tilde{\mathbf{p}}^e \times)\tilde{\mathbf{C}}_b^e \delta\boldsymbol{\omega}_{ib}^b \\
&= \mathbf{d}\overline{\mathbf{v}}_r - (\boldsymbol{\omega}_{ie}^e \times)\mathbf{dp}_r - (\tilde{\mathbf{p}}^e \times)\tilde{\mathbf{C}}_b^e \delta\boldsymbol{\omega}_{ib}^b
\end{aligned} \quad (16)$$

For the inertial sensors, if we only consider the drift bias and noise, $\delta\boldsymbol{\omega}_{ib}^b$ and $\delta\mathbf{f}^b$ can be given by

$$\delta\boldsymbol{\omega}_{ib}^b = \boldsymbol{\varepsilon}^b + \boldsymbol{\eta}_g^b \quad (17a)$$

$$\delta\mathbf{f}^b = \nabla^b + \boldsymbol{\eta}_a^b \quad (17b)$$

where $\boldsymbol{\varepsilon}^b$ is gyroscope drift bias and $\boldsymbol{\eta}_g^b$ the corresponding noise. $\nabla^b$ is accelerometer drift bias and $\boldsymbol{\eta}_a^b$ the corresponding noise.

Define the state vector as

$$\mathbf{dx}_r = \begin{bmatrix} \boldsymbol{\varphi}_r & \mathbf{d}\overline{\mathbf{v}}_r & \mathbf{dp}_r & \boldsymbol{\varepsilon}^b & \nabla^b \end{bmatrix}^T \quad (18)$$

The corresponding state space model is given by

$$\mathbf{d}\dot{\mathbf{x}}_r = \mathbf{F}_r \mathbf{dx}_r + \mathbf{G}_r \boldsymbol{\eta}^b \quad (19a)$$

where

$$\mathbf{F}_r = \begin{bmatrix}
-(\boldsymbol{\omega}_{ie}^e \times) & \mathbf{0}_{3\times 3} & \mathbf{0}_{3\times 3} & -\tilde{\mathbf{C}}_b^e & \mathbf{0}_{3\times 3} \\
(\overline{\mathbf{g}}^e \times) & -(\boldsymbol{\omega}_{ie}^e \times) & \mathbf{0}_{3\times 3} & -(\tilde{\mathbf{v}}^e \times)\tilde{\mathbf{C}}_b^e & -\tilde{\mathbf{C}}_b^e \\
\mathbf{0}_{3\times 3} & \mathbf{I}_{3\times 3} & -(\boldsymbol{\omega}_{ie}^e \times) & -(\tilde{\mathbf{p}}^e \times)\tilde{\mathbf{C}}_b^e & \mathbf{0}_{3\times 3} \\
\mathbf{0}_{3\times 3} & \mathbf{0}_{3\times 3} & \mathbf{0}_{3\times 3} & \mathbf{0}_{3\times 3} & \mathbf{0}_{3\times 3} \\
\mathbf{0}_{3\times 3} & \mathbf{0}_{3\times 3} & \mathbf{0}_{3\times 3} & \mathbf{0}_{3\times 3} & \mathbf{0}_{3\times 3}
\end{bmatrix} \quad (19b)$$

$$\mathbf{G}_r = \begin{bmatrix}
-\tilde{\mathbf{C}}_b^e & \mathbf{0}_{3\times 3} \\
-(\tilde{\mathbf{v}}^e \times)\tilde{\mathbf{C}}_b^e & -\tilde{\mathbf{C}}_b^e \\
-(\tilde{\mathbf{p}}^e \times)\tilde{\mathbf{C}}_b^e & \mathbf{0}_{3\times 3} \\
\mathbf{0}_{6\times 3} & \mathbf{0}_{6\times 3}
\end{bmatrix} \quad (19c)$$

$\boldsymbol{\eta}^b = [\boldsymbol{\eta}_g^{bT} \ \boldsymbol{\eta}_a^{bT}]^T$. It is shown that if we do not consider inertial sensors' error, the state transition matrix is independent of the global state. This is just striking result of the *group affine* property. The trajectory independent linearized model on one hand can handle the filtering inconsistency caused by false-observability of certain state element. This has been widely investigated in field of SLAM and VINS. On the other hand, it opens the door to *log-linearity* in Kalman filtering. The *log-linearity* is not the typical Jacobian linearization along a trajectory because the nonlinear error on the group can be exactly recovered from its solution. Such conclusion is very profitable for attitude alignment with arbitrary misalignments [34-36].

***Remark 3:*** Although incorporating the drift biases into the error state will cause state-dependence for the resultant augmented state transition matrix, much of the benefit provided by the *log-linearity* can be reserved [28]. That is to say, it is still possible to obtain performance improvement making use of the model (19) compared with traditional error state model. Moreover, in inertial based integration, the indirect procedure is always used for attitude, velocity and position estimation. However, for inertial sensors errors, they can be incorporated directly into state and estimated in an open-loop procedure. That is to say, the estimate of these inertial sensors errors will not be feed back into the global state propagation. With such procedure, the dependence of the inertial sensors errors' model on the global state estimate will not affect the filtering performance.

***Remark 4:*** The same state-space model as (19) has been derived in [31]. However, in [31], the state-space model is used in INS/GPS integration. As will be shown in Section IV that, for INS/GPS integration, the left error state model, derived in the next subsection, will be more preferred. This is because that for the group state (5), the values provided by GPS are in left invariant form.

*B. Left Error State Model*

According to (A15), the left group state error is defined as

$$\boldsymbol{\eta}_l = \tilde{\boldsymbol{\chi}}^{-1}\boldsymbol{\chi} = \begin{bmatrix} \tilde{\mathbf{C}}_b^{eT}\mathbf{C}_b^e & \tilde{\mathbf{C}}_b^{eT}\left(\bar{\mathbf{v}}^e - \tilde{\bar{\mathbf{v}}}^e\right) & \tilde{\mathbf{C}}_b^{eT}\left(\mathbf{p}^e - \tilde{\mathbf{p}}^e\right) \\ \mathbf{0}_{1\times 3} & 1 & 0 \\ \mathbf{0}_{1\times 3} & 0 & 1 \end{bmatrix} \quad (20)$$

Denote the attitude error corresponding to $\tilde{\mathbf{C}}_b^{eT}\mathbf{C}_b^e$ in Euler angle form as $\boldsymbol{\varphi}_l$. If $\boldsymbol{\varphi}_l$ assumed to be small value, its relationship with $\tilde{\mathbf{C}}_b^{eT}\mathbf{C}_b^e$ is given by

$$\tilde{\mathbf{C}}_b^{eT}\mathbf{C}_b^e \approx \mathbf{I}_{3\times 3} + \left(\boldsymbol{\varphi}_l \times\right) \quad (21)$$

The differential equation of $\boldsymbol{\varphi}_l$ is given by

$$\dot{\boldsymbol{\varphi}}_l = -\left(\tilde{\boldsymbol{\omega}}_{ib}^b \times\right)\boldsymbol{\varphi}_l - \delta\boldsymbol{\omega}_{ib}^b \quad (22)$$

With approximations (21) and (A16), the vector-form velocity and position errors corresponding to (20) can be derived as

$$\mathbf{d}\bar{\mathbf{v}}_l \approx \tilde{\mathbf{C}}_b^{eT}\left(\bar{\mathbf{v}}^e - \tilde{\bar{\mathbf{v}}}^e\right) = -\tilde{\mathbf{C}}_b^{eT}\delta\bar{\mathbf{v}}^e \quad (23)$$

$$\mathbf{d}\mathbf{p}_l \approx \tilde{\mathbf{C}}_b^{eT}\left(\mathbf{p}^e - \tilde{\mathbf{p}}^e\right) = -\tilde{\mathbf{C}}_b^{eT}\delta\mathbf{p}^e \quad (24)$$

Taking differential operation on both sides of (23) gives

$$\mathbf{d}\dot{\bar{\mathbf{v}}}_l = -\tilde{\mathbf{C}}_b^{eT}\delta\dot{\bar{\mathbf{v}}}^e - \dot{\tilde{\mathbf{C}}}_b^{eT}\delta\bar{\mathbf{v}}^e \quad (25)$$

Before deriving the detailed form of (25), the differential equation of $\delta\bar{\mathbf{v}}^e$ with the attitude error definition in (21) should be firstly derived as

$$\delta\dot{\bar{\mathbf{v}}}^e = \tilde{\mathbf{C}}_b^e\left(\tilde{\mathbf{f}}^b \times\right)\boldsymbol{\varphi}_l - \left(\boldsymbol{\omega}_{ie}^e \times\right)\delta\bar{\mathbf{v}}^e + \tilde{\mathbf{C}}_b^e\delta\mathbf{f}^b \quad (26)$$

Substituting (4) and (26) into (25) gives

$$\begin{aligned}\mathbf{d}\dot{\bar{\mathbf{v}}}_l &= -\left(\tilde{\mathbf{f}}^b \times\right)\boldsymbol{\varphi}_l + \tilde{\mathbf{C}}_b^{eT}\left(\boldsymbol{\omega}_{ie}^e \times\right)\delta\bar{\mathbf{v}}^e - \delta\mathbf{f}^b \\ &\quad + \left(\tilde{\boldsymbol{\omega}}_{ib}^b \times\right)\tilde{\mathbf{C}}_b^{eT}\delta\bar{\mathbf{v}}^e - \tilde{\mathbf{C}}_b^{eT}\left(\boldsymbol{\omega}_{ie}^e \times\right)\delta\bar{\mathbf{v}}^e \\ &= -\left(\tilde{\mathbf{f}}^b \times\right)\boldsymbol{\varphi}_l - \left(\tilde{\boldsymbol{\omega}}_{ib}^b \times\right)\mathbf{d}\bar{\mathbf{v}}_l - \delta\mathbf{f}^b\end{aligned} \quad (27)$$

Taking differential operation on both sides of (24) gives

$$\mathbf{d}\dot{\mathbf{p}}_l = -\tilde{\mathbf{C}}_b^{eT}\delta\dot{\mathbf{p}}^e - \dot{\tilde{\mathbf{C}}}_b^{eT}\delta\mathbf{p}^e \quad (28)$$

Substituting (4) and (12) into (28) gives

$$\mathbf{d}\dot{\mathbf{p}}_l = \mathbf{d}\bar{\mathbf{v}}_l + (\tilde{\boldsymbol{\omega}}_{ib}^b \times)\tilde{\mathbf{C}}_b^{eT}\delta\mathbf{p}^e = \mathbf{d}\bar{\mathbf{v}}_l - (\tilde{\boldsymbol{\omega}}_{ib}^b \times)\mathbf{d}\mathbf{p}_l \quad (29)$$

Define the state vector as

$$\mathbf{dx}_l = \begin{bmatrix}\boldsymbol{\varphi}_l & \mathbf{d}\bar{\mathbf{v}}_l & \mathbf{d}\mathbf{p}_l & \boldsymbol{\varepsilon}^b & \nabla^b\end{bmatrix}^T \quad (30)$$

The corresponding state space model is given by

$$\mathbf{d}\dot{\mathbf{x}}_l = \mathbf{F}_l \mathbf{dx}_l + \mathbf{G}_l \boldsymbol{\eta}^b \quad (31a)$$

where

$$\mathbf{F}_l = \begin{bmatrix} -\left(\tilde{\boldsymbol{\omega}}_{ib}^b \times\right) & \mathbf{0}_{3\times 3} & \mathbf{0}_{3\times 3} & -\mathbf{I}_{3\times 3} & \mathbf{0}_{3\times 3} \\ -\left(\tilde{\mathbf{f}}^b \times\right) & -\left(\tilde{\boldsymbol{\omega}}_{ib}^b \times\right) & \mathbf{0}_{3\times 3} & \mathbf{0}_{3\times 3} & -\mathbf{I}_{3\times 3} \\ \mathbf{0}_{3\times 3} & \mathbf{I}_{3\times 3} & -\left(\tilde{\boldsymbol{\omega}}_{ib}^b \times\right) & \mathbf{0}_{3\times 3} & \mathbf{0}_{3\times 3} \\ \mathbf{0}_{3\times 3} & \mathbf{0}_{3\times 3} & \mathbf{0}_{3\times 3} & \mathbf{0}_{3\times 3} & \mathbf{0}_{3\times 3} \\ \mathbf{0}_{3\times 3} & \mathbf{0}_{3\times 3} & \mathbf{0}_{3\times 3} & \mathbf{0}_{3\times 3} & \mathbf{0}_{3\times 3} \end{bmatrix} \quad (31b)$$

$$\mathbf{G}_l = \begin{bmatrix} -\mathbf{I}_{3\times 3} & \mathbf{0}_{3\times 3} \\ \mathbf{0}_{3\times 3} & -\mathbf{I}_{3\times 3} \\ \mathbf{0}_{9\times 3} & \mathbf{0}_{9\times 3} \end{bmatrix} \quad (31c)$$

Interestingly, for the left error definition, the state transition matrix is still independent of the global state even when incorporating the inertial sensors' errors into the state.

## IV. APPLICATION ASPECTS

### A. Error State Model Selection

It is known from last section that there are two different error state definitions and corresponding error state models. How to select the error state model in practical application can be determined according to the invariant type of the observations. The invariant observations are defined as [23, 28]

$$\text{Left-Invariant Observation: } \mathbf{y} = \boldsymbol{\chi}\mathbf{b} \quad (32a)$$

$$\text{Rright-Invariant Observation: } \mathbf{y} = \boldsymbol{\chi}^{-1}\mathbf{b} \quad (32b)$$

where $\mathbf{b}$ is a constant vector. If one observation satisfies (32), the resultant linearized observation model will also be autonomous. More specifically, if the observation is left-invariant, with the left error definition, the linearized observation model will be independent of the global state. Similarly, if the observation is right-invariant, with the right error definition, the linearized observation model will be independent of the global state.

In this section, we consider two typical applications, that is, INS/GPS integration and INS/Odometer integration. For INS/GPS integration, the GPS can provide accurate velocity and position. Therefore, the observation can be given by

$$\mathbf{y}_{\text{GPS}} = \mathbf{v}_{\text{GPS}}^e + \left(\boldsymbol{\omega}_{ie}^e \times\right)\mathbf{p}_{\text{GPS}}^e \quad (33)$$

It can be easily verified that the observation in (33) is a left-invariant observation with the group state (5). In this paper, the linearized state-space model is used in the indirect integration. With the left error definition (30), the observation transition function is given by

$$\mathbf{z}_{\text{obser}} = \underbrace{\begin{bmatrix}\mathbf{0}_{3\times 3} & -\tilde{\mathbf{C}}_b^e & \mathbf{0}_{3\times 9}\end{bmatrix}}_{\mathbf{H}_l}\mathbf{dx}_l \quad (34)$$

where

$$\mathbf{z}_{\text{obser}} = \tilde{\bar{\mathbf{v}}}^e - \mathbf{y}_{\text{GPS}} \quad (35)$$

The transition matrix in (34) is seemingly dependent on the global state $\tilde{\mathbf{C}}_b^e$. However, as pointed out in [22] that "*Applying a linear function to the innovation term of an EKF before computing the gains does not change the results of the filter*". Therefore, we can construct a transformed observation and the corresponding transition matrix as

$$\bar{\mathbf{z}}_{\text{obser}} = \tilde{\mathbf{C}}_b^{eT}\mathbf{z}_{\text{obser}} \quad (36)$$

$$\bar{\mathbf{H}}_l = \tilde{\mathbf{C}}_b^{eT}\mathbf{H}_l = \begin{bmatrix}\mathbf{0}_{3\times 3} & -\mathbf{I}_{3\times 3} & \mathbf{0}_{3\times 9}\end{bmatrix} \quad (37)$$

Now, it is clearly shown that $\bar{\mathbf{H}}_l$ is no longer dependent on the global state.

Similarly, with the right error definition (18), the observation transition function corresponding to (33) is given by

$$\mathbf{z}_{\text{obser}} = \underbrace{\begin{bmatrix}\left(\tilde{\bar{\mathbf{v}}}^e \times\right) & \mathbf{I}_{3\times 3} & \mathbf{0}_{3\times 9}\end{bmatrix}}_{\mathbf{H}_r}\mathbf{dx}_l \quad (38)$$

It is clearly shown that the observation transition $\mathbf{H}_r$ is dependent on the global state $\tilde{\bar{\mathbf{v}}}^e$. When such model is used in Kalman filtering, performance may be degraded if $\tilde{\bar{\mathbf{v}}}^e$ has not been well estimated. However, there will be no such problem for model (34). In this respect, the left error state model will be more preferred for INS/GPS integration.

For INS/Odometer integration, the observation is the body velocity $\mathbf{v}^b$. With the group state (5), it can be verified that $\mathbf{v}^b$ is neither left-invariant nor right-invariant. However, if we consider the non-transformed group state

$$\boldsymbol{\gamma} = \begin{bmatrix} \mathbf{C}_b^e & \mathbf{v}^e & \mathbf{p}^e \\ \mathbf{0}_{1\times 3} & 1 & 0 \\ \mathbf{0}_{1\times 3} & 0 & 1 \end{bmatrix} \quad (39)$$

It can be verified that

$$\mathbf{v}^b = \boldsymbol{\gamma}^{-1}\mathbf{b} = \begin{bmatrix} \mathbf{C}_b^{eT} & -\mathbf{C}_b^{eT}\mathbf{v}^e & -\mathbf{C}_b^{eT}\mathbf{p}^e \\ \mathbf{0}_{1\times 3} & 1 & 0 \\ \mathbf{0}_{1\times 3} & 0 & 1 \end{bmatrix}\begin{bmatrix} 0 \\ -1 \\ 0 \end{bmatrix} \quad (40)$$

That is to say, $\mathbf{v}^b$ is right-invariant for the group state (39). In this respect, we can say that $\mathbf{v}^b$ is *more like* right-invariant for the group state (5). Therefore, it is still appropriate to apply the right error definition. With right error definition (18), the indirect observation is given by

$$\begin{aligned}
\mathbf{z} &= \tilde{\mathbf{v}}^e - (\boldsymbol{\omega}_{ie}^e \times)\tilde{\mathbf{p}}^e - \tilde{\mathbf{C}}_b^e \mathbf{v}^b = \tilde{\mathbf{v}}^e - \tilde{\mathbf{C}}_b^e \mathbf{v}^b = -(\mathbf{v}^e \times)\boldsymbol{\varphi}_r + \boldsymbol{\delta v}^e \\
&= -\left[\overline{\mathbf{v}}^e - (\boldsymbol{\omega}_{ie}^e \times)\mathbf{p}^e\right] \times \boldsymbol{\varphi}_r + \boldsymbol{\delta \overline{v}}^e - (\boldsymbol{\omega}_{ie}^e \times)\boldsymbol{\delta p}^e \\
&\approx \left[-(\tilde{\mathbf{p}}^e \times)(\boldsymbol{\omega}_{ie}^e \times)\right]\boldsymbol{\varphi}_r - \mathbf{d}\overline{\mathbf{v}}_r + (\boldsymbol{\omega}_{ie}^e \times)\mathbf{dp}_r
\end{aligned} \quad (41)$$

It is shown that the transition matrix derived from (41) is dependent on the global state $\tilde{\mathbf{p}}^e$.

Similarly, with the left error definition (30), the indirect observation is given by

$$\begin{aligned}
\mathbf{z} &= \tilde{\mathbf{v}}^e - (\boldsymbol{\omega}_{ie}^e \times)\tilde{\mathbf{p}}^e - \tilde{\mathbf{C}}_b^e \mathbf{v}^b = \tilde{\mathbf{v}}^e - \tilde{\mathbf{C}}_b^e \mathbf{v}^b = -(\mathbf{v}^e \times)\tilde{\mathbf{C}}_b^e \boldsymbol{\varphi}_l + \boldsymbol{\delta v}^e \\
&= -\left[\overline{\mathbf{v}}^e - (\boldsymbol{\omega}_{ie}^e \times)\mathbf{p}^e\right] \times \tilde{\mathbf{C}}_b^e \boldsymbol{\varphi}_l + \boldsymbol{\delta \overline{v}}^e - (\boldsymbol{\omega}_{ie}^e \times)\boldsymbol{\delta p}^e \\
&\approx \left[-(\tilde{\tilde{\mathbf{v}}}^e \times) + (\boldsymbol{\omega}_{ie}^e \times)(\tilde{\mathbf{p}}^e \times) - (\tilde{\mathbf{p}}^e \times)(\boldsymbol{\omega}_{ie}^e \times)\right]\tilde{\mathbf{C}}_b^e \boldsymbol{\varphi}_l \\
&\quad - \tilde{\mathbf{C}}_b^e \mathbf{d}\overline{\mathbf{v}}_l + (\boldsymbol{\omega}_{ie}^e \times)\tilde{\mathbf{C}}_b^e \mathbf{dp}_l
\end{aligned} \quad (42)$$

It is clearly shown that with left error definition, the transition matrix derived from (42) depends on $\tilde{\mathbf{C}}_b^e$, $\tilde{\tilde{\mathbf{v}}}^e$ and $\tilde{\mathbf{p}}^e$. In other words, the global state-dependence of (42) is more severe than that of (41). Especially, the global attitude dependence in (42) may introduce much negative effect on the filtering performance, because the attitude is always more difficult to be estimated than velocity and position. In this respect, the right error state model will be more preferred for INS/Odometer integration.

### B. Initial Covariance Setting

The initial covariance setting can affect the filtering performance. Generally, if the state vector is not correlated with each other, the diagonal elements of the covariance can be given empirically according to the pre-knowledge degree of these values. However, for the correlated state defined in this paper, the covariance cannot be set so directly based on the experience. If we have set an initial covariance $\mathbf{P}_{avp,0}$ for the error state $[\boldsymbol{\varphi}^T \; \boldsymbol{\delta v}^{eT} \; \boldsymbol{\delta p}^{eT}]^T$ empirically, the initial covariance $\mathbf{P}_{avp,r0}$ for the right error state $[\boldsymbol{\varphi}_r^T \; \mathbf{d\overline{v}}_r^T \; \mathbf{dp}_r^T]^T$ can be determined according to the transformation (3), (10) and (11) as

$$\mathbf{P}_{avp,r0} = \mathbf{T}_R \mathbf{P}_{avp,0} \mathbf{T}_R^T \quad (43)$$

where

$$\mathbf{T}_R = \begin{bmatrix} \mathbf{I}_{3\times 3} & \mathbf{0}_{3\times 3} & \mathbf{0}_{3\times 3} \\ (\tilde{\overline{\mathbf{v}}}_0^e \times) & -\mathbf{I}_{3\times 3} & -(\boldsymbol{\omega}_{ie}^e \times) \\ (\tilde{\mathbf{p}}_0^e \times) & \mathbf{0}_{3\times 3} & -\mathbf{I}_{3\times 3} \end{bmatrix} \quad (44)$$

where $\tilde{\overline{\mathbf{v}}}_0^e$ is the initial velocity and $\tilde{\mathbf{p}}_0^e$ is the initial position.

The initial covariance $\mathbf{P}_{avp,l0}$ for the left error state $[\boldsymbol{\varphi}_l^T \; \mathbf{d\overline{v}}_l^T \; \mathbf{dp}_l^T]^T$ can be determined according to the transformation (3), (23) and (24) as

$$\mathbf{P}_{avp,l0} = \mathbf{T}_L \mathbf{P}_{avp,0} \mathbf{T}_L^T \quad (45)$$

where

$$\mathbf{T}_L = \begin{bmatrix} \mathbf{I}_{3\times 3} & \mathbf{0}_{3\times 3} & \mathbf{0}_{3\times 3} \\ \mathbf{0}_{3\times 3} & -\tilde{\mathbf{C}}_{b,0}^{e\ T} & -\tilde{\mathbf{C}}_{b,0}^{e\ T}(\boldsymbol{\omega}_{ie}^e \times) \\ \mathbf{0}_{3\times 3} & \mathbf{0}_{3\times 3} & -\tilde{\mathbf{C}}_{b,0}^{e\ T} \end{bmatrix} \quad (46)$$

where $\tilde{\mathbf{C}}_{b,0}^e$ is the initial attitude matrix.

*Remark 5:* The initial covariance transformation is very crucial for inertial based integration, especially for the case where the position is of interest. For INS/Odometer integration, if the initial covariance is setting without transformation as shown in (44), the position estimate may even not converge.

### C. Feedback Correction

There are three sequential stages for the indirect integration at one time instant, i.e. INS calculation, Kalman filtering and feedback correction. The Kalman filtering is virtually used to estimate the error state as defined in (18) or (30). The estimated error state by Kalman filtering is used to refine the INS calculation results, which is just the feedback correction. The feedback correction is corresponding to the error state definition and can be viewed as the inverse operation. For the right error definition, the feedback correction at time instant $k$ is given by

$$\hat{\mathbf{C}}_{b,k}^n = \exp(\hat{\boldsymbol{\varphi}}_{r,k})\tilde{\mathbf{C}}_{b,k}^n \quad (47a)$$

$$\hat{\tilde{\mathbf{v}}}_k^e = \tilde{\overline{\mathbf{v}}}_k^e + \mathbf{d}\hat{\mathbf{v}}_{r,k} - \tilde{\overline{\mathbf{v}}}_k^e \times \hat{\boldsymbol{\varphi}}_{r,k} \quad (47b)$$

$$\hat{\mathbf{p}}_k^e = \tilde{\mathbf{p}}_k^e + \mathbf{d}\hat{\mathbf{p}}_{r,k} - \tilde{\mathbf{p}}_k^e \times \hat{\boldsymbol{\varphi}}_{r,k} \quad (47c)$$

where $\tilde{\mathbf{C}}_{b,k}^n$, $\tilde{\overline{\mathbf{v}}}_k^e$ and $\tilde{\mathbf{p}}_k^e$ are the INS calculation results. $\hat{\boldsymbol{\varphi}}_{r,k}$, $\mathbf{d}\hat{\mathbf{v}}_{r,k}$ and $\mathbf{d}\hat{\mathbf{p}}_{r,k}$ are the error state estimate by the Kalman filtering. $\hat{\mathbf{C}}_{b,k}^n$, $\hat{\tilde{\mathbf{v}}}_k^e$ and $\hat{\mathbf{p}}_k^e$ are the refined navigation parameters and will be used as inputs for INS to calculate $\tilde{\mathbf{C}}_{b,k+1}^n$, $\tilde{\overline{\mathbf{v}}}_{k+1}^e$ and $\tilde{\mathbf{p}}_{k+1}^e$ at next time instant.

Similarly, for the left error definition, the feedback correction at time instant $k$ is given by

$$\hat{\mathbf{C}}_{b,k}^e = \tilde{\mathbf{C}}_{b,k}^e \exp(\hat{\boldsymbol{\varphi}}_{l,k}) \quad (48a)$$

$$\hat{\tilde{\mathbf{v}}}_k^e = \tilde{\mathbf{C}}_{b,k}^e \mathbf{d}\hat{\mathbf{v}}_{l,k} + \tilde{\overline{\mathbf{v}}}_k^e \quad (48b)$$

$$\hat{\mathbf{p}}_k^e = \tilde{\mathbf{C}}_{b,k}^e \mathbf{d}\hat{\mathbf{p}}_{l,k} + \tilde{\mathbf{p}}_k^e \quad (48c)$$

After the state update, the corresponding error state estimate should be reset to zero, that is $\mathbf{d}\hat{\mathbf{x}}_{r/l,k}(1:9) = \mathbf{0}_{9\times 1}$. (47) and (48) are only used to refine the global state. There is no consensus yet about that is it necessary to update the covariance corresponding to the state update (47) and (48) [37]. In this paper, we adopt the statement that the reset operation does not provide new information and will not perform covariance update for the global state update and reset operations [38].

***Remark 6:*** It should be noted that when performing the correction (47) and (48), the rigid exponential map operation of $\mathbb{SE}_2(3)$ has not been used [39]. The velocity position correction in (47) and (48) can be viewed as linear approximation of the exponential map operation of $\mathbb{SE}_2(3)$. However, such approximation will not lose accuracy. This is because that the exponential map operation is only used to construct the nonlinear group error. However, as has been pointed out that for the group affine dynamic, the nonlinear group error can be recovered accurately from the linear vector error.

## V. EXPERIMENTS STUDY FOR INS/GPS INTEGRATION

In this section, the performance of the right/left error state models is evaluated using simulated and pre-collected field test data for INS/GPS integration. For comparison, the traditional INS mechanization (1) is used for INS calculation. With error state definition

$$\delta \mathbf{x} = \begin{bmatrix} \boldsymbol{\varphi}_r & \delta \mathbf{v}^e & \delta \mathbf{p}^e & \boldsymbol{\varepsilon}^b & \nabla^b \end{bmatrix}^T \quad (49)$$

its corresponding error state model is given by

$$\delta \dot{\mathbf{x}} = \mathbf{F} \delta \mathbf{x} + \mathbf{G} \boldsymbol{\eta}^b \quad (50a)$$

where

$$\mathbf{F} = \begin{bmatrix} -(\boldsymbol{\omega}_{ie}^e \times) & \mathbf{0}_{3\times 3} & \mathbf{0}_{3\times 3} & -\tilde{\mathbf{C}}_b^e & \mathbf{0}_{3\times 3} \\ (\tilde{\mathbf{C}}_b^e \tilde{\mathbf{f}}^b \times) & -2(\boldsymbol{\omega}_{ie}^e \times) & \mathbf{0}_{3\times 3} & \mathbf{0}_{3\times 3} & \tilde{\mathbf{C}}_b^e \\ \mathbf{0}_{3\times 3} & \mathbf{I}_{3\times 3} & \mathbf{0}_{3\times 3} & \mathbf{0}_{3\times 3} & \mathbf{0}_{3\times 3} \\ \mathbf{0}_{3\times 3} & \mathbf{0}_{3\times 3} & \mathbf{0}_{3\times 3} & \mathbf{0}_{3\times 3} & \mathbf{0}_{3\times 3} \\ \mathbf{0}_{3\times 3} & \mathbf{0}_{3\times 3} & \mathbf{0}_{3\times 3} & \mathbf{0}_{3\times 3} & \mathbf{0}_{3\times 3} \end{bmatrix} \quad (50b)$$

$$\mathbf{G} = \begin{bmatrix} -\tilde{\mathbf{C}}_b^e & \mathbf{0}_{3\times 3} \\ \mathbf{0}_{3\times 3} & \tilde{\mathbf{C}}_b^e \\ \mathbf{0}_{9\times 3} & \mathbf{0}_{9\times 3} \end{bmatrix} \quad (50c)$$

The indirect measurement provided by GPS is given by

$$\mathbf{z}_{obser} = \begin{bmatrix} \tilde{\mathbf{v}}^e - \mathbf{v}_{GPS}^e \\ \tilde{\mathbf{p}}^e - \mathbf{p}_{GPS}^e \end{bmatrix} = \begin{bmatrix} \mathbf{0}_{3\times 3} & \mathbf{I}_{3\times 3} & \mathbf{0}_{3\times 3} & \mathbf{0}_{3\times 6} \\ \mathbf{0}_{3\times 3} & \mathbf{0}_{3\times 3} & \mathbf{I}_{3\times 3} & \mathbf{0}_{3\times 6} \end{bmatrix} \delta \mathbf{x} \quad (51)$$

Specially, the following three integration algorithms are evaluated and compared. The integration algorithm making use of INS calculation model (4) and left error model (31) is denoted as LSE-KF. The integration algorithm making use of INS calculation model (4) and right error model (19) is denoted as RSE-KF. The integration algorithm making use of INS calculation model (1) and error model (50) is denoted as SO-KF.

### A. Simulation Results

This section conducts simulations to evaluate the performance of different error state-space models for INS/GPS integration. The vehicle is equipped with a navigation-grade IMU, which includes a triad of gyroscopes (bias $0.01°/h$, noise $0.001°/\sqrt{h}$) and accelerometers (bias $100\mu g$, noise $10\mu g/\sqrt{Hz}$). The initial true attitude is set as $[0\ 0\ 0]°$ and the vehicle is under static condition during the simulation. The simulation time is 300s and the IMU update interval T = 0.01s. For the three evaluated algorithms, the initial yaw error covariance is set as $(160°)^2$ and the initial pitch and roll error covariance are both set as $(60°)^2$. The initial attitude estimates for the three algorithms are generated randomly. Specially, for pith and roll, the initial estimate is generated randomly according to $\mathcal{N}(0, (60°)^2)$ and for yaw the initial estimate is generated randomly according to $\mathcal{N}(0, (160°)^2)$. A total of 200 Monte Carlo runs are conducted for the three algorithms. The attitude estimate errors by the three algorithms are shown in Fig. 1-3, respectively.

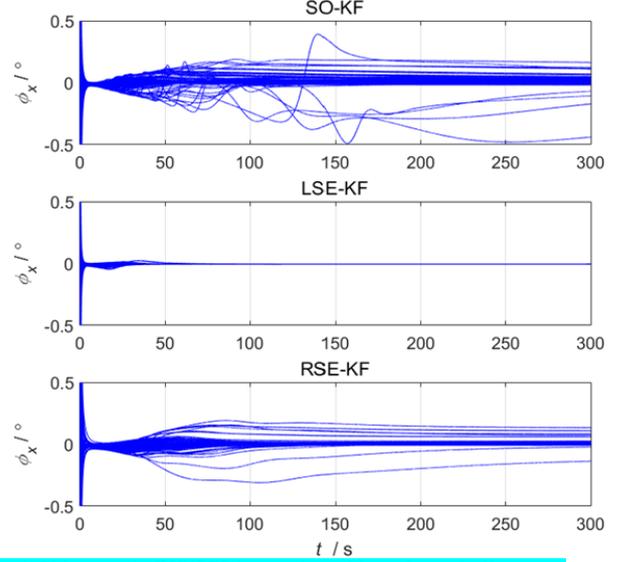

Fig. 1. Pith errors in INS/GPS integration with random misalignments

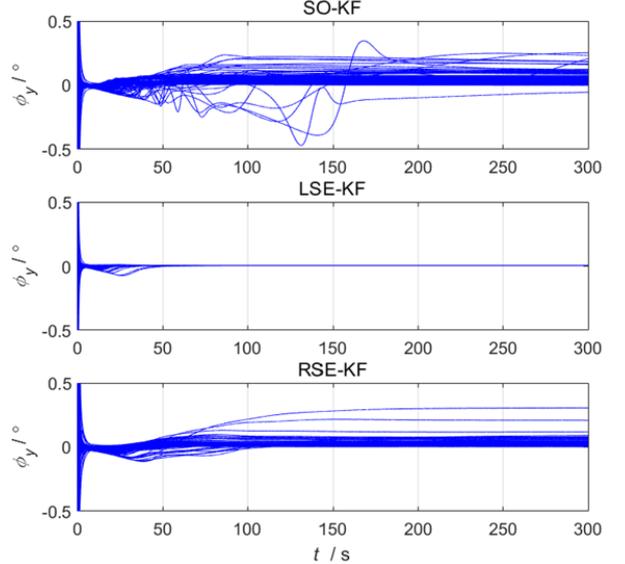

Fig. 2. Roll errors in INS/GPS integration with random misalignments

It is clearly shown that the LSE-KF performs best among the three algorithms, which is consistent with the theoretical analysis. This is because that there is no global state estimate in the error state-space model for LSE-KF and therefore, the inaccurate estimate of these global state will not affect the filtering performance. In contrast, the error state-space models are dependent on the global state estimate for both RSE-KF and SO-KF. Specially, for RSE-KF such dependence inherent in the measurement model and the involved global state estimate is the velocity. In contrast, for SO-KF such dependence inherent in the process model and the involved global state estimate is the attitude matrix. It is well known that for INS/GPS integration, the attitude is more difficult to be estimated than velocity. Therefore, the dependence of attitude matrix will introduce more negative effect on the filtering performance than that by velocity. This is just the reason why RSE-KF performs better than SO-KF.

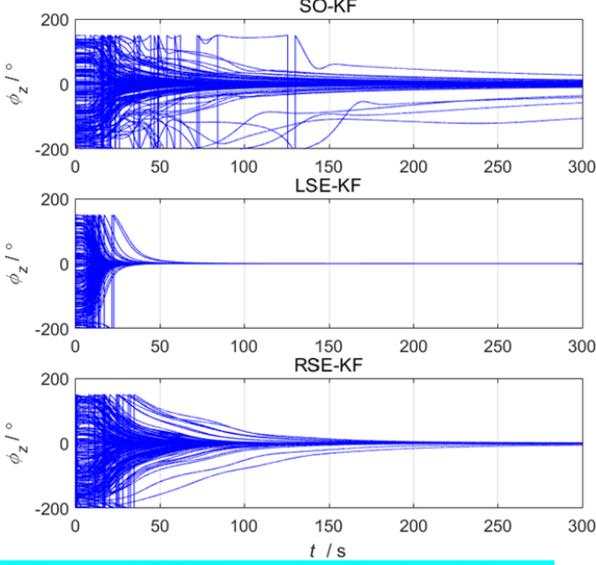

Fig. 3. Yaw errors in INS/GPS integration with random misalignment

### B. Field Test Results

A car-mounted test is carried out for algorithms evaluation. In this experiment, the reference attitude is provided by a POS, which is equipped with a triad of gyroscopes (drift $0.01°/h$) and accelerometers (bias $20\mu g$). The data of an intermediate-grade fiber optic INS with a triad of gyroscopes (drift $0.3°/h$) and accelerometers (bias $20mg$) is collected for algorithms evaluation. The INS sampling rate is 200Hz and the GPS sampling rate is 1Hz. The velocity and position provided by GPS are used as measurements. The length of the collected data is about 800s and the test trajectory is shown in Fig. 4.

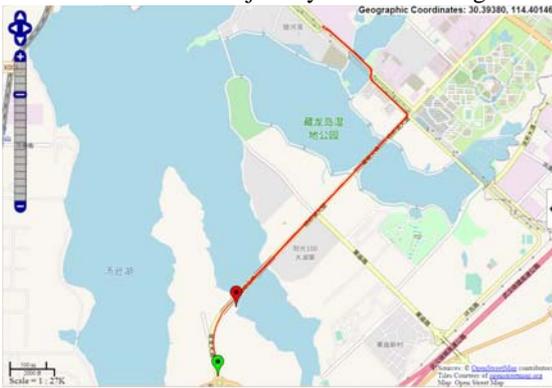

Fig. 4. Test trajectory for INS/GPS integration

Firstly, the initial attitude error is set as $[1° \ 1° \ 3°]^T$ and the initial velocity and position are provided by GPS directly. For the three evaluated algorithms, the initial yaw error covariance is set as $(3°)^2$ and the initial pitch and roll error covariance are both set as $(1°)^2$. The attitude results by the three algorithms are shown in Fig. 5-7, respectively. It is shown that there is no obvious difference between the three algorithms. In [21], it is pointed out that making use of the global state independent model can cope with the inconsistency problem caused by false-observability in field of SLAM and VINS. However, for INS/GPS integration, such inconsistency problem is not so obvious, because the heading angle in INS/GPS integration can also be observable. The non-observable state may be the inertial sensors' drift bias. However, for all the three evaluated algorithms, the drift bias has not been feedback into the INS calculation. That is to say, the drift bias is estimated in an open-loop manner and the attitude, velocity and position are estimated in close-loop manner. Therefore, the inertial sensors' drift bias will not affect the attitude, velocity and position estimate. The reason for open-loop estimation of inertial sensors' drift bias is that the inertial sensors' drift bias cannot be estimated so well within short time period.

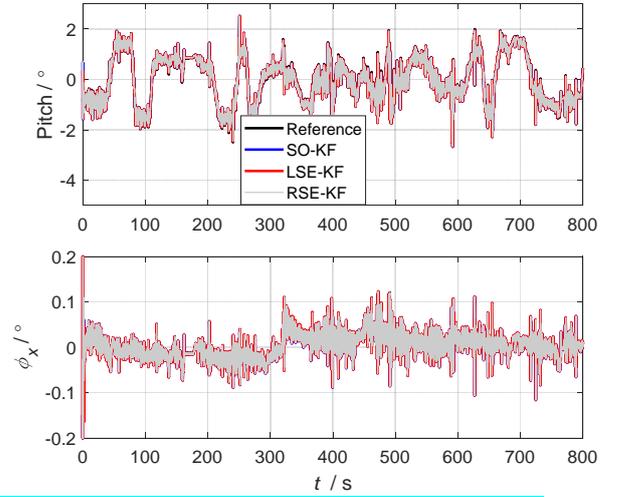

Fig. 5. Pith results in INS/GPS integration with small misalignment

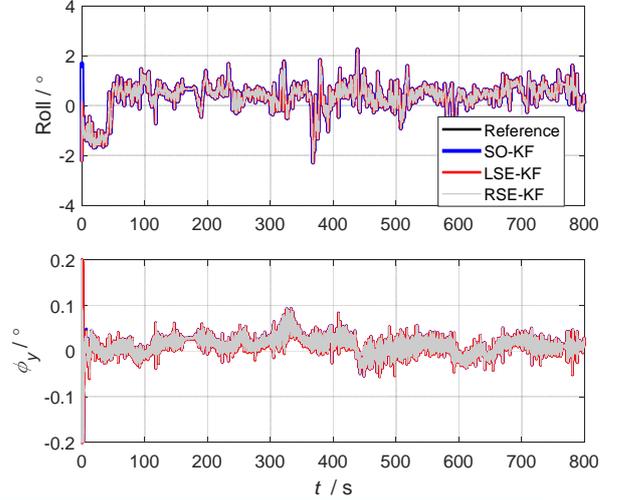

Fig. 6. Roll results in INS/GPS integration with small misalignment

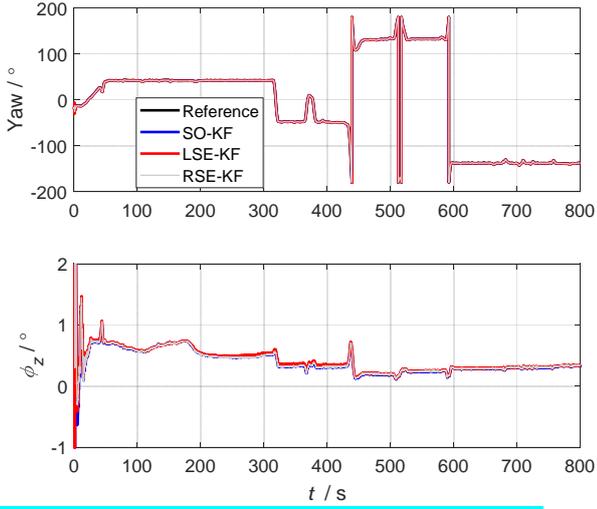
Fig. 7. Yaw results in INS/GPS integration with small misalignment

According to the above test results analysis, we can know that the advantage of the global state-independent model has not been exploited since the false-observability in INS/GPS integration is not so obvious. However, another advantage of the left error model (31) is that the nonlinear group error $\boldsymbol{\eta}_l$ can be exactly recovered from the linear error $[\boldsymbol{\varphi}_l^T \ \mathbf{d\bar{v}}_l^T \ \mathbf{dp}_l^T]^T$. In other word, the linear model (31) can also be applied in case with very large initial attitude error, which has been shown in the simulation results. With this consideration, the initial attitude error is set as $[60° \ 60° \ 160°]^T$. For the three evaluated algorithms, the initial yaw error covariance is set as $(160°)^2$ and the initial pitch and roll error covariance are both set as $(60°)^2$. The attitude estimate errors by the three algorithms are shown in Fig.8-10, respectively. For clarity, only the first 100s is presented for the three attitude angles. It is shown that all the three algorithms can converge although their models are derived with assumed small attitude misalignments. However, LSE-KF and RSE-KF converge faster than SO-KF and their steady-state errors are also smaller than SO-KF. These results are consistent with the simulation results. The corresponding reasons have also been discussed in the simulation study. Meanwhile, LSE-KF also outperforms RSE-KF due to the fact that its error state-space model is totally global state-independent. This conclusion is consistent with that in [40].

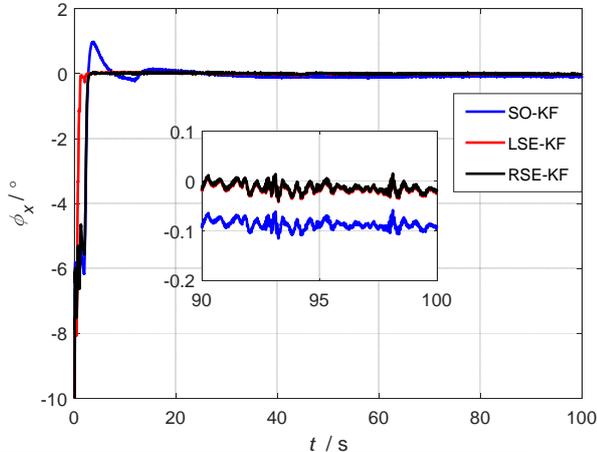
Fig. 8. Pith errors in INS/GPS integration with large misalignment

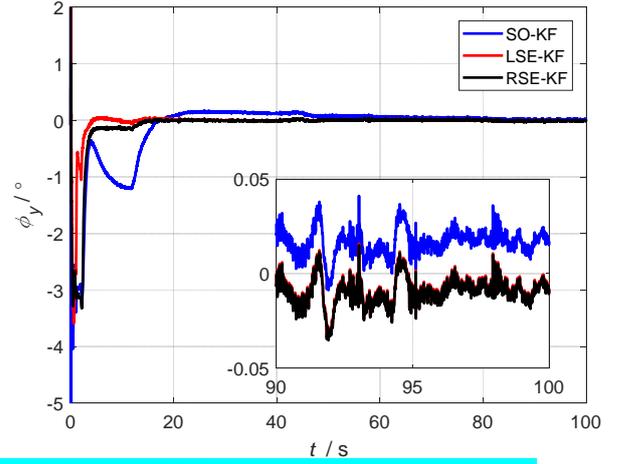
Fig. 9. Roll errors in INS/GPS integration with large misalignment

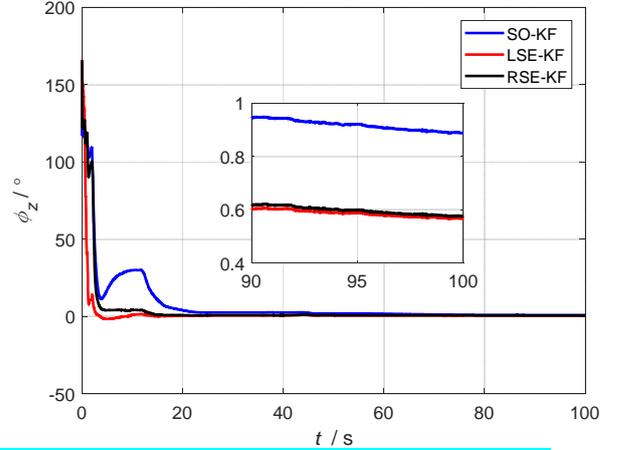
Fig. 10. Yaw errors in INS/GPS integration with large misalignment

## VI. EXPERIMENTS STUDY FOR INS/ ODOMETER INTEGRATION

### A. Simulation Results

This section conducts simulations to evaluate the performance of different error state space models for INS/ Odometer integration. The simulation conditions are all the same with those in Section V.A. Since the vehicle is under static condition, the indirect observations are given by $\mathbf{y} = \tilde{\mathbf{v}}^e - (\boldsymbol{\omega}_{ie}^e \times)\tilde{\mathbf{p}}^e$ for SO-KF and $\mathbf{y} = \tilde{\tilde{\mathbf{v}}}^e - (\boldsymbol{\omega}_{ie}^e \times)\tilde{\mathbf{p}}^e$ for LSE-KF and RSE-KF. With error state definition (49), the measurement model for SO-KF is given by

$$\begin{aligned} \mathbf{z} &= \tilde{\tilde{\mathbf{v}}}^e - (\boldsymbol{\omega}_{ie}^e \times)\tilde{\mathbf{p}}^e - \tilde{\mathbf{C}}_b^e \mathbf{v}^b \\ &= \tilde{\mathbf{v}}^e - \tilde{\mathbf{C}}_b^e \mathbf{v}^b \approx \underbrace{\begin{bmatrix} -(\tilde{\mathbf{v}}^e \times) & \mathbf{I}_{3\times 3} & \mathbf{0}_{3\times 9} \end{bmatrix}}_{\mathbf{H}} \delta\mathbf{x} \end{aligned} \quad (52)$$

The attitude estimate errors across 200 Monte Carlo runs by the three algorithms are shown in Fig. 11-13, respectively. It is clearly shown that for INS/Odometer integration, RSE-KF performs best, which is consistent with the theoretical analysis. In INS/Odometer integration, the attitude is also more difficult to estimate compared with the velocity and position. For LSE-KF, the measurement transition matrix is dependent on the global attitude matrix and for SO-KF the process transition matrix is dependent on the global attitude matrix. So their performance has been degraded due to the dependence of the global attitude estimate. From Fig. 11-13, it cannot be definitely judged whether SO-KF or LSE-KF performs better. That is to say, it cannot be definitely judged which form dependence will introduce more negative effect in filtering performance, dependence in process model or in measurement model. Meanwhile, through comparison between Fig. 1-3 and Fig. 11-13, it can also be observed that the INS attitude can be estimated more accurate aided by GPS than that aided by Odometer. This is because that the attitude observability of INS/GPS is stronger than that of INS/Odometer.

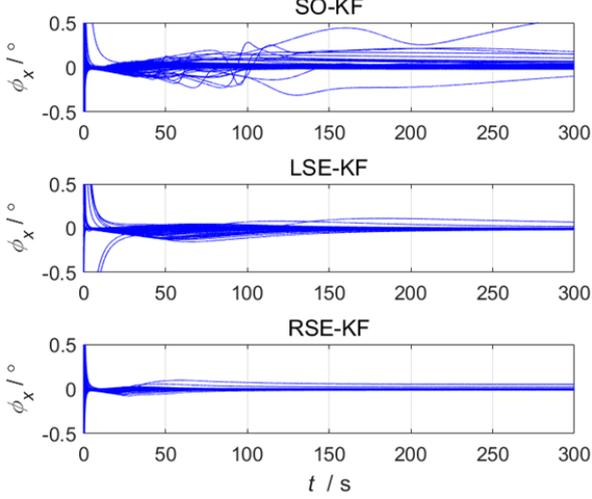

Fig. 11. Pith errors in INS/ Odometer integration with random misalignments

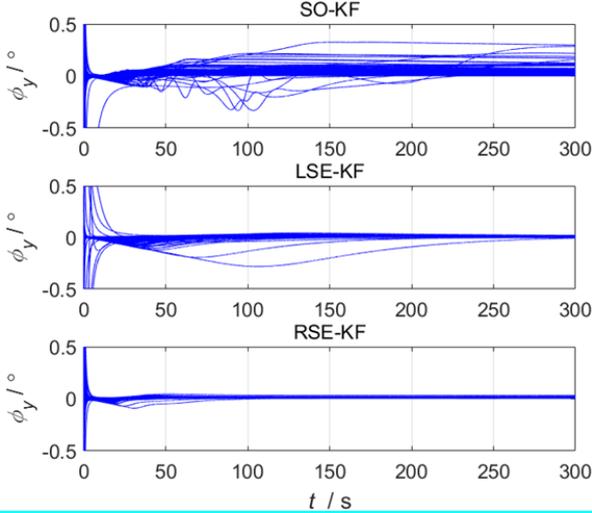

Fig. 12. Roll errors in INS/ Odometer integration with random misalignments

Another important performance index of INS/Odometer integration is its long-endurance position accuracy. With this consideration, a complex dynamic simulation scenario is carried out. The trajectory is generated according to the settings listed in Table I. In Table I, wx denotes the pith angular rate, wy roll angular rate and wz yaw angular rate, respectively. Similarly, ax denotes lateral acceleration, ay longitudinal acceleration and az up acceleration, respectively. The reference attitude, velocity and position are generated making use of these motion parameters. The generated angular rate and specific fore are then added the gyroscopes errors (bias $0.01°/h$, noise $0.001°/\sqrt{h}$) and accelerometers errors (bias $100\mu g$, noise $10\mu g/\sqrt{Hz}$) to simulate the IMU measurements. The total time of the simulation is 18880 seconds. The simulated trajectory is shown in Fig.14.

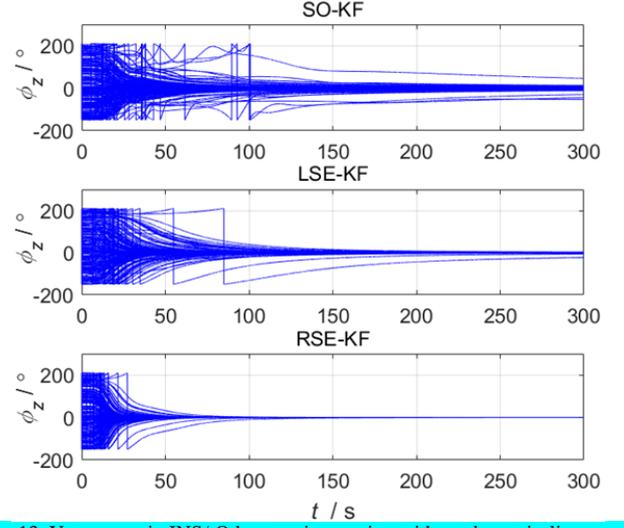

Fig. 13. Yaw errors in INS/ Odometer integration with random misalignments

TABLE I. MOTION PARAMETERS FOR SIMULATED DATA GENERATION
(ACC-accelerate, CS-constant speed, LT-left turn, RT-right turn, DEC-decelerate, w=0.9°/s, a=9)

| Status | Duration | wx | wy | wz | ax | ay | az |
|---|---|---|---|---|---|---|---|
| static | 100s | 0 | 0 | 0 | 0 | 0 | 0 |
| ACC | 10s | 0 | 0 | 0 | 0 | 1 | 0 |
| CS | 3000s | 0 | 0 | 0 | 0 | 0 | 0 |
| LT | 100s | 0 | 0 | w | -a | 0 | 0 |
| CS | 3000s | 0 | 0 | 0 | 0 | 0 | 0 |
| RT | 100s | 0 | 0 | -w | a | 0 | 0 |
| CS | 1000s | 0 | 0 | 0 | 0 | 0 | 0 |
| RT | 100s | 0 | 0 | -w | a | 0 | 0 |
| CS | 1000s | 0 | 0 | 0 | 0 | 0 | 0 |
| RT | 100s | 0 | 0 | -w | a | 0 | 0 |
| CS | 5000s | 0 | 0 | 0 | 0 | 0 | 0 |
| LT | 100s | 0 | 0 | w | -a | 0 | 0 |
| CS | 3000s | 0 | 0 | 0 | 0 | 0 | 0 |
| LT | 100s | 0 | 0 | w | -a | 0 | 0 |
| CS | 1000s | 0 | 0 | 0 | 0 | 0 | 0 |
| LT | 100s | 0 | 0 | w | -a | 0 | 0 |
| CS | 1000s | 0 | 0 | 0 | 0 | 0 | 0 |
| DEC | 10s | 0 | 0 | 0 | 0 | -1 | 0 |
| static | 60 | 0 | 0 | 0 | 0 | 0 | 0 |

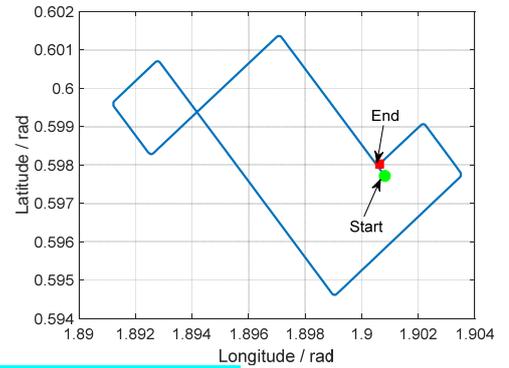

Fig. 14. Simulated dynamic trajectory

For the three evaluated algorithms, the initial attitude misalignment is set as $[0.5\ -0.5\ 30]^T$ arcmin. The initial velocity error is set as 0.1 m/s and the initial latitude and longitude errors are both set as 10 m. For the three evaluated algorithms, the initial yaw error covariance is set as $(1°)^2$ and the initial pitch and roll error covariance are both set as $(0.1°)^2$. The latitude and longitude estimate errors by the three algorithms are shown in Fig. 15 and the attitude errors are shown in Fig. 16. It is clearly shown that the RSE-KF performs much better than the other two algorithms. After the vehicle accelerating to the constant speed, the yaw angle has not been correctly estimated by LSE-KF and SO-KF. So their position errors have been accumulated much faster than RSE-KF. In this simulation scenario, LSE-KF performs better than SO-KF, which means that the negative effect by the global attitude-dependence in process model (50b) is more severe than that by the global attitude dependence in process model (42). From the results, it can be observed that the frequent maneuvering has a much negative effect on the traditional integration filter, that is SO-KF, for INS/Odometer. In practical applications with severe maneuvering, data processing should be delicately performed for odometer measurements before they can be used in the integration filer. In contrast, the RSE-KF is natural immune to these frequent maneuvering. Therefore, the RSE-KF would be more preferred in complex application environments.

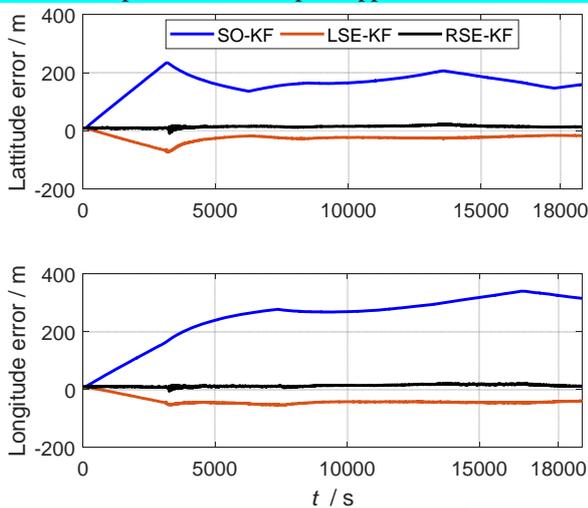

Fig. 15. Latitude and longitude errors by the three algorithms

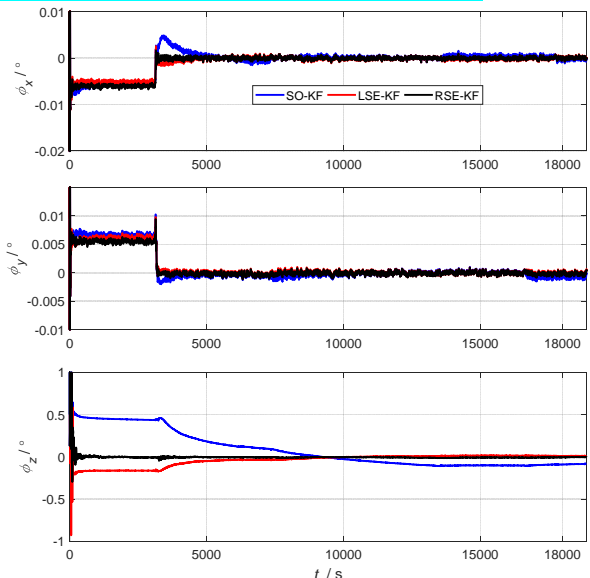

Fig. 16. Attitude errors by the three algorithms

### B. Field Test Results

In this section, the INS/Odometer integration is carried out making use of field test data. The test data was collected from a car-mounted experiment. On the experiment platform there is a navigation-grade INS which contains three ring laser gyroscopes with drift of $0.007°/h$ and three quartz accelerometers with bias of $30\mu g$. Besides the INS, the experiment platform also contains a GPS receiver and Odometer. The GPS can provide velocity with precision of about $0.1m/s$ and position with precision of about 10m at frequency $1Hz$. The Odometer can provide body velocity with precision of about ±0.5% of its speed at frequency $125Hz$. The results of INS/GPS integration are used as reference. The experiment trajectory is shown in Fig. 17 and the field tests last about 3500s with the distance about $46km$.

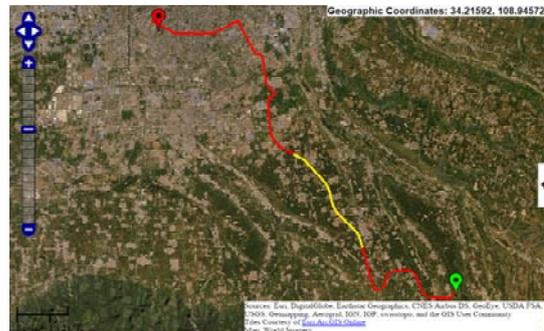

Fig. 17. Field tests with the distance about $46km$

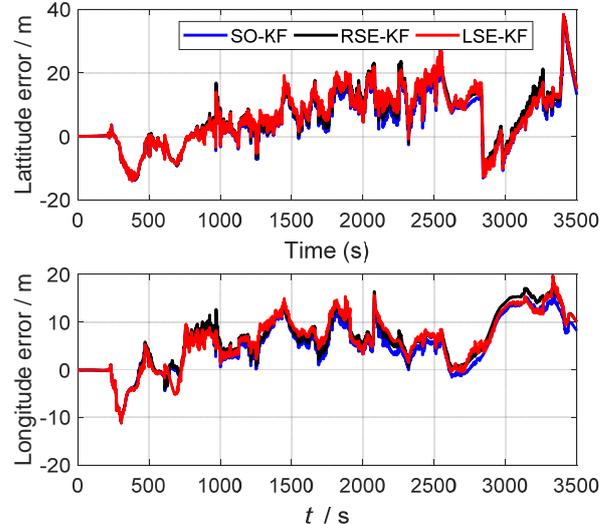

Fig. 18. Position estimate errors in INS/Odometer integration

For the three evaluated algorithms, the initial attitude, velocity and position are all set with the reference values. The initial yaw error covariance is set as $(1°)^2$ and the initial pitch and roll error covariance are both set as $(0.1°)^2$. The latitude and longitude errors by the three algorithms are shown in Fig. 18. It is shown that the three algorithms perform quit similar with each other. The attitude estimate results by the three algorithms are shown in Fig. 19-21, respectively. Based on the observability analysis, the position estimate accuracy of INS/Odometer integration is mainly affected by the yaw angle estimate accuracy [41, 42]. As can be seen from the yaw angels estimate results in Fig. 21 that, there are frequent movements in yaw direction, which can improve the corresponding estimate accuracy. The attitude results by the three algorithms are also very similar with each other, which is consistent with the position results.

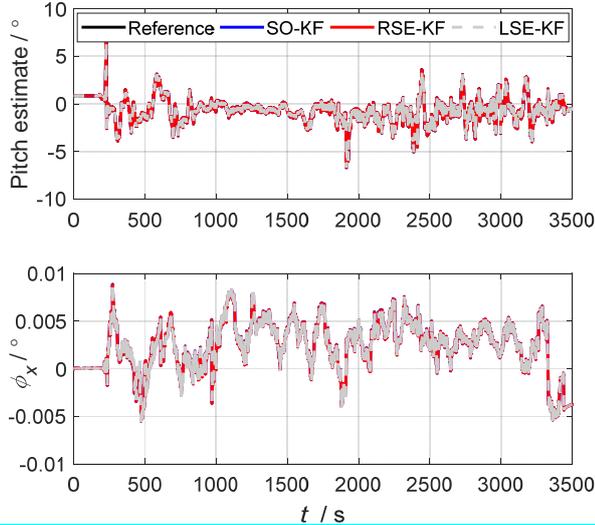

Fig. 19. Pitch results in INS/Odometer integration with small misalignment

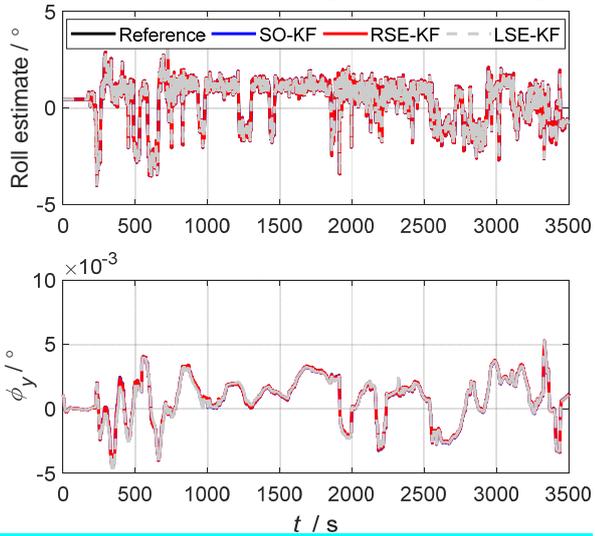

Fig. 20. Roll results in INS/Odometer integration with small misalignment

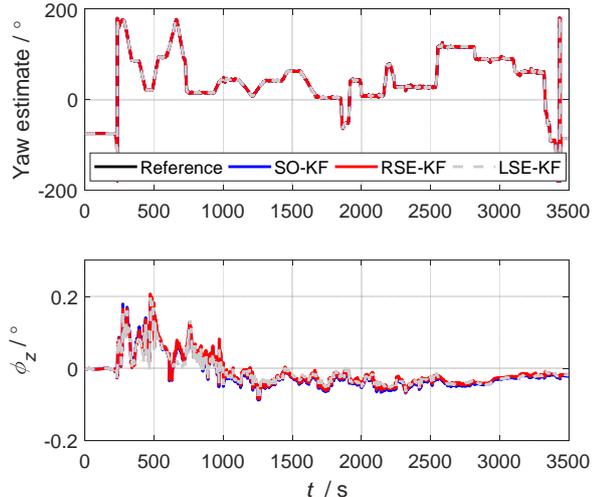

Fig. 21. Yaw results in INS/Odometer integration with small misalignment

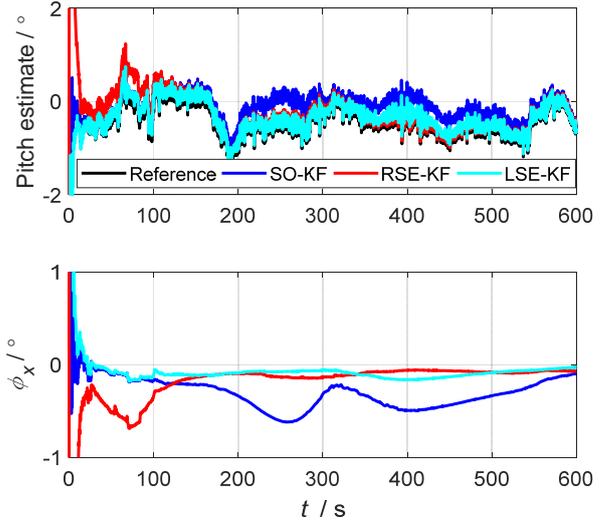

Fig. 22. Pitch results in INS/Odometer integration with large misalignment

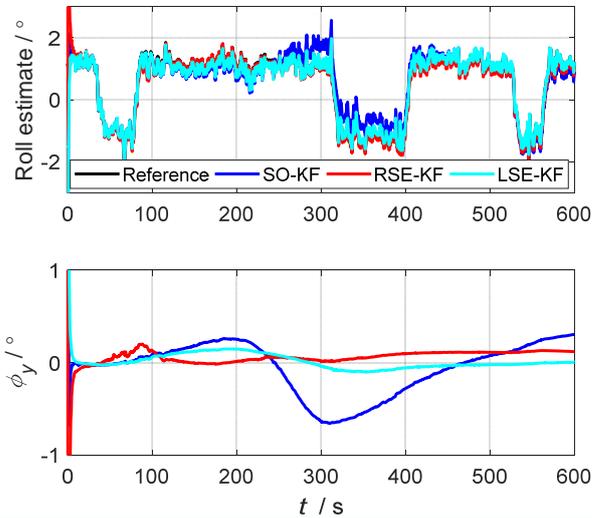

Fig. 23. Roll results in INS/Odometer integration with large misalignment

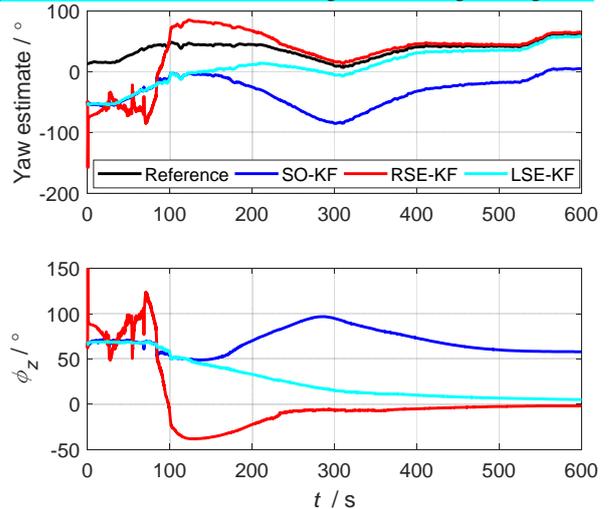

Fig. 24. Yaw results in INS/Odometer integration with large misalignment

Another advantage of the $\mathbb{SE}_2(3)$ based model is its ability to handle the large initial misalignment due to its log-linearity. Therefore, we would also like to check this striking property for INS/Odometer integration. With this consideration, a 600s dynamic data segment is used to evaluate the two models with large initial misalignments. The trajectory of the selected data segment is marked with yellow color in Fig. 17. The initial attitude error is set as $[30° \ 30° \ 60°]^T$. The initial yaw error covariance is set as $(60°)^2$ and the initial pitch and roll error covariance are both set as $(30°)^2$ for the three evaluated algorithms. The initial position is set directly using the reference position. The initial velocity is assumed to be zero, which is because that only the body frame velocity can be measured in this application situation. The attitude results by the three algorithms are shown in Fig.22-24, respectively. It is shown that both RSE-KF and LSE-KF can outperform SO-KF. For yaw angle estimate, RSE-KF performs best. While for pith and roll angles estimate, LSE-KF performs even a litter better than RSE-KF. However, such superiority is not so obvious. The ability of handling the large initial misalignments by RSE-KF and LSE-KF is caused by the log-linearity of the attitude, velocity and position error equations in (19b) and (31b). The degraded performance of SO-KF is due to its process models' dependence on the global attitude estimate. The position errors during the alignment by the three algorithms are shown in Fig.25. It is clearly shown that the RSE-KF performs best. It is known that the accumulated position error is mainly caused by yaw errors. Therefore, the results in Fig. 25 is just corresponding to those in Fig. 24. However, with the right error definition, the measurement transition matrix is not totally trajectory independent, as shown in (41). Therefore, the state-space model (19) and (41) cannot handle extreme large initial misalignments. Actually, we have also carried out the test with initial attitude error $[60° \ 60° \ 160°]^T$ and all the three algorithms cannot converge.

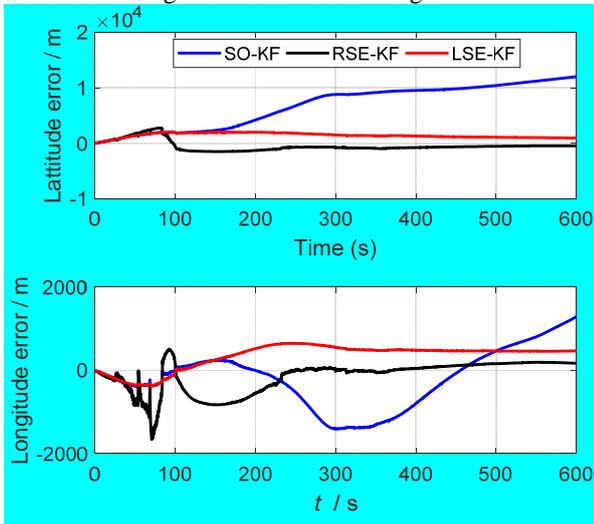

Fig. 25. Position errors during the alignment

## VII. CONCLUSION

In this paper, the error state models are derived corresponding to a transformed INS mechanization in ECEF. Since the transformed INS mechanization satisfies a *group affine* property with group state on $\mathbb{SE}_2(3)$, the resultant right and left error state models are trajectory-independent. Then the error state models are applied in INS/GPS and INS/Odometer integration. It is shown that the advantage of remedying filtering inconsistency by the trajectory-independent error state models has not been brought into play for the two applications. This is because that the false-observability problem of certain state element is not obvious in the two applications. However, another advantage of the trajectory-independent error state models is their log-linearity, which is very suitable for attitude initialization with arbitrary misalignments. For INS/GPS integration, with left error definition, both the process model and measurement model are trajectory-independent. Therefore, such linear state-space model can still work quit well with even extreme large misalignments. For INS/Odometer integration, however, the measurement model is not totally trajectory independent. The corresponding state-space model can only relax the initial attitude accuracy requirement to a certain extent. Moreover, for long-endurance INS/Odometer integration, the right error state-space model will be more preferred compared with traditional model due to its natural immune ability to severe maneuvering. The extensive simulation and experimental studies in this paper also indicates that the $\mathbb{SE}_2(3)$ formulation is not essentially superior over $\mathbb{SO}(3)$. How to select the state formulation is dependent on the applications scenarios.

## APPENDIX

### MATRIX LIE GROUP USEFUL FORMULAS

In this section, some useful matrix Lie group formulas are presented, which can be used to derive the corresponding error state models. In this paper, the matrix Lie group is only used to formulate different nonlinear group errors and then to derive the vector errors definition. So, the presented formulas are only some used equations in this paper. To familiarize with Lie group theory, one is suggested to refer to [43].

The special orthogonal group $\mathbb{SO}(3)$ is used to the represent the rotation matrix $\mathbf{C}$ as

$$\mathbb{SO}(3) = \left\{ \mathbf{C} \in \mathbb{SO}(3) \middle| \mathbf{CC}^T = \mathbf{I}_{3\times 3}, \det \mathbf{C} = 1 \right\} \quad (A1)$$

The Lie algebra associated with $\mathbb{SO}(3)$ is given by

$$\mathfrak{so}(3) = \left\{ \mathbf{\Phi} = (\mathbf{\varphi}\times) \in \mathbb{R}^{3\times 3} \middle| \mathbf{\varphi} \in \mathbb{R}^3 \right\} \quad (A2)$$

where

$$(\mathbf{\varphi}\times) = \begin{bmatrix} \varphi_1 \\ \varphi_2 \\ \varphi_3 \end{bmatrix} \times = \begin{bmatrix} 0 & -\varphi_3 & \varphi_2 \\ \varphi_3 & 0 & -\varphi_1 \\ -\varphi_2 & \varphi_1 & 0 \end{bmatrix} \quad (A3)$$

The exponential map is the key to relating a matrix Lie group to its associated Lie algebra. For $\mathbb{SO}(3)$, such relation is given by

$$\exp(\mathbf{\varphi}\times) = \exp(\varphi \mathbf{a}\times) = \underbrace{\cos\varphi \mathbf{I}_{3\times 3} + (1-\cos\varphi)\mathbf{a}\mathbf{a}^T + \sin\varphi(\mathbf{a}\times)}_{\mathbf{C}} \quad (A4)$$

The inverse operation of (A4) is given by

$$\varphi = \cos^{-1}\left((\operatorname{tr}(\mathbf{C})-1)/2\right), \mathbf{C}\mathbf{a} = \mathbf{a} \quad (A5)$$

The group of double direct spatial isometries $\mathbb{SE}_2(3)$ is used to represent the extended pose as

$$\mathbb{SE}_2(3) = \left\{ \mathbf{T} = \begin{bmatrix} \mathbf{C} & \mathbf{v} & \mathbf{p} \\ \mathbf{0}_{2\times 3} & \mathbf{I}_{2\times 2} \end{bmatrix} \in \mathbb{R}^{5\times 5} \middle| \begin{array}{c} \mathbf{C} \in \mathbb{SO}(3) \\ \mathbf{v}, \mathbf{p} \in \mathbb{R}^3 \end{array} \right\} \quad (A6)$$

The Lie algebra associated with $\mathbb{SE}_2(3)$ is given by

$$\mathfrak{se}_2(3) = \left\{ \mathbf{\Xi} = \boldsymbol{\zeta}^\wedge \in \mathbb{R}^{5\times 5} \middle| \boldsymbol{\zeta} \in \mathbb{R}^9 \right\} \quad (A7)$$

where

$$\boldsymbol{\zeta}^\wedge = \begin{bmatrix} \boldsymbol{\varphi} \\ \boldsymbol{\upsilon} \\ \boldsymbol{\rho} \end{bmatrix}^\wedge = \begin{bmatrix} \boldsymbol{\varphi}\times & \boldsymbol{\upsilon} & \boldsymbol{\rho} \\ \mathbf{0}_{2\times 3} & \mathbf{0}_{2\times 2} \end{bmatrix} \in \mathbb{R}^{5\times 5}, \ \boldsymbol{\upsilon}, \boldsymbol{\rho} \in \mathbb{R}^3 \quad (A8)$$

The exponential map relating $\mathbb{SE}_2(3)$ and $\mathfrak{se}_2(3)$ is given by

$$\exp(\boldsymbol{\zeta}^\wedge) = \begin{bmatrix} \exp(\boldsymbol{\varphi}\times) & \mathbf{J}_l \boldsymbol{\upsilon} & \mathbf{J}_l \boldsymbol{\rho} \\ \mathbf{0}_{2\times 3} & \mathbf{I}_{2\times 2} \end{bmatrix} \quad (A9)$$

where

$$\mathbf{J}_l = \frac{\sin\varphi}{\varphi}\mathbf{I}_{3\times 3} + \left(1 - \frac{\sin\varphi}{\varphi}\right)\mathbf{a}\mathbf{a}^T + \left(\frac{1-\cos\varphi}{\varphi}\right)(\mathbf{a}\times) \quad (A10)$$

The inverse operation of (A9) is given by

$$\boldsymbol{\zeta} = \ln(\mathbf{T})^\vee = \begin{bmatrix} \varphi\mathbf{a} \\ \mathbf{J}_l^{-1}\mathbf{v} \\ \mathbf{J}_l^{-1}\mathbf{p} \end{bmatrix} \quad (A11)$$

where $\mathbf{T}$ is with the form in (A6) and

$$\mathbf{J}_l^{-1} = \frac{\varphi}{2}\cot\frac{\varphi}{2}\mathbf{I}_{3\times 3} + \left(1 - \frac{\varphi}{2}\cot\frac{\varphi}{2}\right)\mathbf{a}\mathbf{a}^T - \frac{\varphi}{2}(\mathbf{a}\times) \quad (A12)$$

In this paper, the vector error state is derived corresponding to the group error. In order to define the group error, the inverse of the group state is necessary. For $\mathbb{SE}_2(3)$, its inverse is given by

$$\mathbf{T}^{-1} = \begin{bmatrix} \mathbf{C}^T & -\mathbf{C}^T\mathbf{v} & -\mathbf{C}^T\mathbf{p} \\ \mathbf{0}_{2\times 3} & \mathbf{I}_{2\times 2} \end{bmatrix} \in \mathbb{SE}_2(3) \quad (A13)$$

It is shown that the inverse of $\mathbb{SE}_2(3)$ is also $\mathbb{SE}_2(3)$.

For $\mathbb{SE}_2(3)$, denote one realization of $\mathbf{T}$ as $\bar{\mathbf{T}}$. The right group error is defined as

$$\delta\mathbf{T}_{\text{right}} = \mathbf{T}\bar{\mathbf{T}}^{-1} \ or \ \bar{\mathbf{T}}\mathbf{T}^{-1} \quad (A14)$$

The left group error is defined as

$$\delta\mathbf{T}_{\text{left}} = \bar{\mathbf{T}}^{-1}\mathbf{T} \ or \ \mathbf{T}^{-1}\bar{\mathbf{T}} \quad (A15)$$

When deriving the linear vector error state model, the attitude error is always assumed to be small. In this case, $\mathbf{J}_l^{-1}$ in (A11) can be approximated as

$$\mathbf{J}_l^{-1} \approx \mathbf{I}_{3\times 3} - (\boldsymbol{\varphi}\times)/2 \quad (A16)$$

Denote the group error $\delta\mathbf{T}$ as

$$\delta\mathbf{T} = \begin{bmatrix} \delta\mathbf{C} & \delta\mathbf{v} & \delta\mathbf{p} \\ \mathbf{0}_{2\times 3} & \mathbf{I}_{2\times 2} \end{bmatrix} \quad (A17)$$

With the approximation (A16), the vector error corresponding to $\delta\mathbf{T}$ can be approximated according to (A11) as

$$\boldsymbol{\zeta} \approx \begin{bmatrix} \varphi\mathbf{a}^T & \delta\mathbf{v}^T & \delta\mathbf{p}^T \end{bmatrix}^T \quad (A18)$$

The relation between $\delta\mathbf{C}$ and $\varphi\mathbf{a}$ is given by (A5).

ACKNOWLEDGMENT

The authors would like to thank Prof. Gongmin Yan from Northwestern Polytechnical University for providing the INS/Odometer integration test data.

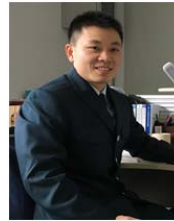

**Lubin Chang** received the B.Sc. and Ph.D. degrees in navigation from the Department of Navigation Engineering, Naval University of Engineering, Wuhan, China in 2009 and 2014, respectively.

He is currently working in Naval University of Engineering as an Associate Professor. He has published more than 30 scientific papers in international journals. His research interests include inertial navigation systems, inertial-based integrated navigation systems, gravity aided navigation, and state estimation theory. Prof. Chang has been awarded a fellowship from the Alexander von Humboldt Foundation of Germany in 2016. He was the recipient of the 2015 Excellent Doctoral Dissertation of Hubei Provinces and the 2016 National Postdoctoral Program for Innovative Talents. He is the winner of the 2014 IET Premium Best Paper Award in Science, Measurement and Technology. He has been an Associate Editor of the *IEEE Access* and an Editorial Board Member of the *IET Radar, Sonar & Navigation* since 2017.

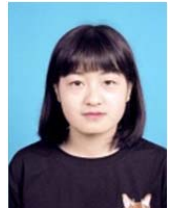

**Jingbo Di** received the B.Sc. degree From Liaoning Institute of Science and Technology in 2020. She is currently working toward the M.S. degree in the field of inertial navigation systems with Naval University of Engineering.

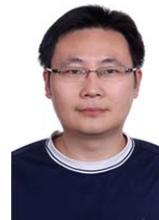

**Fangjun Qin** received the B.Sc., M.Sc. and Ph.D. degrees in navigation from the Department of Navigation, Naval University of Engineering, Wuhan, P.R. China in 2002, 2005 and 2009 respectively.

Currently, he is a professor of the Department of Navigation Engineering with the Naval University of Engineering. His scientific interests include inertial navigation and INS/GPS integrated navigation technology.